\documentclass{article} 
\usepackage{iclr2025,times}

\usepackage{amsmath,amsfonts,bm}

\def\eqref#1{equation~\ref{#1}}

\def\1{\bm{1}}

\DeclareMathAlphabet{\mathsfit}{\encodingdefault}{\sfdefault}{m}{sl}
\SetMathAlphabet{\mathsfit}{bold}{\encodingdefault}{\sfdefault}{bx}{n}

\usepackage{hyperref}
\usepackage{url}
\usepackage{graphicx} 
\usepackage{float}    
\usepackage{booktabs}
\usepackage{multirow}
\usepackage{changepage}
\usepackage{listings}
\usepackage{placeins}
\usepackage{wrapfig}

\title{Performance of Zero-Shot Time Series \\ Foundation Models on Cloud Data}

\author{William Toner\thanks{Equal contribution $^\dagger$Work done while an intern at Huawei} , Thomas~L. Lee$^\ast$$^\dagger$, Artjom Joosen, Rajkarn Singh, Martin Asenov\\
Systems Infrastructure Lab, Edinburgh Research Centre, Huawei \\
\texttt{william.toner2@huawei.com, T.L.Lee-1@sms.ed.ac.uk}
}

\iclrfinalcopy 
\begin{document}

\maketitle

\begin{abstract}
Time series foundation models (FMs) have emerged as a popular paradigm for zero-shot multi-domain forecasting. FMs are trained on numerous diverse datasets and claim to be effective forecasters across multiple different time series domains, including cloud data. In this work we investigate this claim, exploring the effectiveness of FMs on \emph{cloud data}. We demonstrate that many well-known FMs fail to generate meaningful or accurate zero-shot forecasts in this setting. We support this claim empirically, showing that FMs are outperformed consistently by simple linear baselines. We also illustrate a number of interesting pathologies, including instances where FMs suddenly output seemingly erratic, random-looking forecasts. Our results suggest a widespread failure of FMs to model cloud data. 
\end{abstract}
\FloatBarrier
\begin{figure*}[h]
        \centering
        \includegraphics[width=0.99\textwidth]{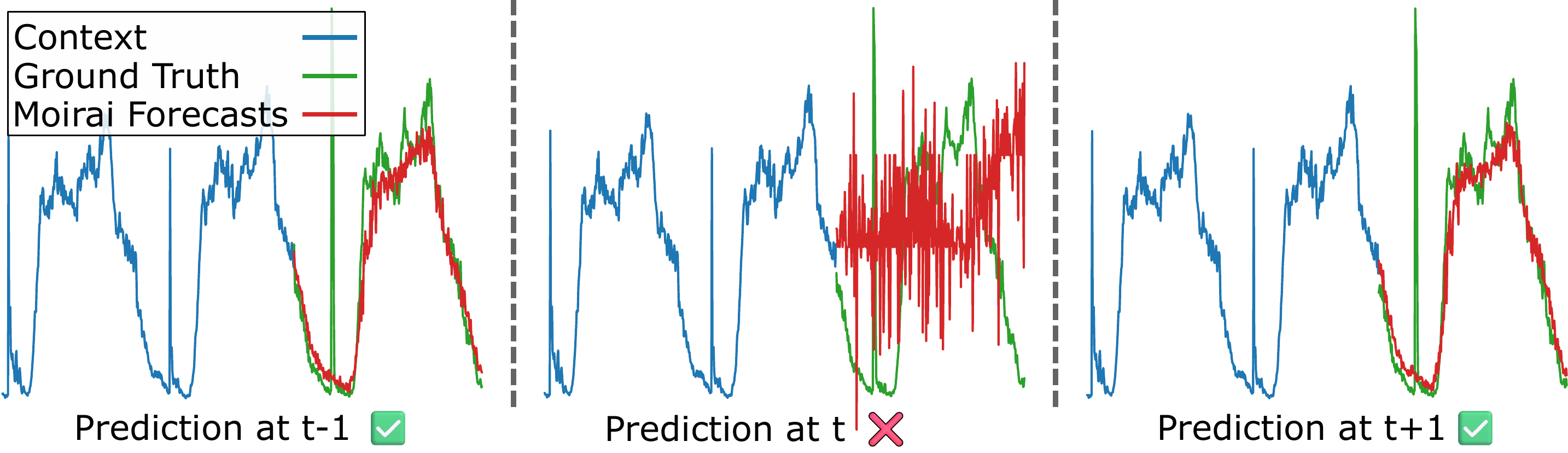}
        \caption{\textbf{Demonstration of the pathological behaviour of zero-shot FMs on cloud data:} The figure shows three consecutive forecasts for the Moirai FM on the Huawei Cloud D2 dataset. The forecasts shown are those produced when starting forecasting at the $t\text{-}1$, $t$, and $t\text{+}1$ time steps, where $t=7130$. The blue curves show the context Moirai is given to construct the forecast. The plot shows that with only a small change in the context, Moirai's forecasts can change from predicting reasonably (the $t\text{-}1$ and $t\text{+}1$\textsuperscript{th} time steps) to giving inaccurate and chaotic forecasts (the $t$\textsuperscript{th} time step).}
        \label{fig:moirai}
\end{figure*}

\section{Introduction}
Recently, there has been a growing trend in developing foundation models for time series forecasting \citep{liang2024foundation, ye2024survey}. \emph{Foundation models} (FMs) are neural networks pretrained on large datasets from diverse time series domains. One of the main motivations of FMs is that they claim to generalise zero-shot (i.e., without training data) to a wide range of time series domains, including cloud data \citep{Woo2024moirai, li2024foundts}. Time series forecasting is important in cloud settings \citep{joosen2023does, diao2024forecasting}. For example, forecasting allows cloud providers to scale up resources pre-emptively and reduce costs associated with over-provisioning \citep{joosen2024serverless}. Furthermore, zero-shot forecasting is needed in cloud settings since it is often the case for cloud data that many time series are created and removed in short periods of time \citep{darlow2024dam}. While the cloud setting is a potentially impactful use-case for FMs, it has been, to the best of our knowledge, rarely evaluated in the FM literature \citep{Ansari2024Chronos, rasul2023lag, chaudhry2019tiny}. In this work we aim to mend this gap by exploring the question of \textit{how well do zero-shot foundation models generalise to cloud data?}

To examine the behaviour of (zero-shot) FMs on cloud time series, we perform experiments on function demand data drawn from real-world usage. Our results show that FMs perform poorly, being outperformed by the simple baselines of a linear model and a naive seasonal forecaster. This draws into question the claim that FMs can generalise to cloud data out-of-the-box. Instead we find that the particular characteristics of cloud data---such as spikeyness \citep{joosen2024serverless, diao2024forecasting}---can make FMs forecast badly. For instance, Figure~\ref{fig:moirai} illustrates the pathological behaviour of the Moirai FM on typical cloud data. The figure demonstrates a minor change in the time series context leading Moirai to give erratic forecasts. We examine the performance of other FMs and find similar irrational forecasts on cloud data, as shown in Figure~\ref{fig:FM_forecasts}. Additionally, we show that the best performing FM in our experiments, VisionTS, performs well due to the fact it gives near-identical forecasts to a naive seasonal forecaster. Overall, our results suggest that current FMs are ineffective for forecasting cloud data.

\section{Related work and Preliminaries}\label{sec:preliminaries}
\textbf{Zero-Shot FMs} \; Recently, many (zero-shot) FM models have been proposed for time series forecasting. Most of these models follow the same general design principles of: \textbf{a)} being a LLM-like transformer model and \textbf{b)} being pretrained on a large corpus of time series data, consisting of billions of training points \citep{rasul2023lag,chen2024visionts,liang2024foundation}. For instance, \textit{Chronos} \citep{Ansari2024Chronos}, \textit{Moirai} \citep{Woo2024moirai} and \textit{TimesFM} \citep{Das2024TimesFM} are FMs which adhere to both of these design principles. However, some newer FMs are designed differently, using non-transformer architectures and/or not being trained on real-world time series data. For example, TTM \citep{Ekambaram2024Tiny}, VisionTS \citep{chen2024visionts} and Mamba4Cast \citep{bhethanabhotlamamba4cast} all use non-transformer architectures. Additionally, Mamba4Cast is trained only on synthetic time series, while VisionTS is not trained on time series at all, using an ImageNet-pretrained masked auto-encoder \citep{he2022masked} as its backbone. 

\textbf{Rolling Window Forecasting} \; Here, we describe the time series setting used in this work. A \emph{time series} consists of a sequence of fixed-dimension vectors $\bm{x}_t$, indexed by the time step $t$. Each dimension of $\bm{x}_t$ is called a \emph{channel}. We look at the realistic scenario of \textit{rolling window} forecasting \citep{nie2022time}, which models the deployment stage of a forecast model. In a rolling window setting, at each time step $t$ the forecaster gives a forecast for $H$ time steps into the future, where $H$ is called the \emph{forecast horizon}. To construct the forecast, the forecaster is given a \emph{context}: a look-back window of the $L$ previous time-series values, $\bm{x}_{t-L}, \ldots, \bm{x}_t$, where we call $L$ the \emph{context length}. To evaluate performance, forecasts are compared against the \emph{ground truth}, $\bm{x}_{t+1}, \ldots, \bm{x}_{t+H}$, i.e. the actual values the time series takes for the forecasted time steps.       

\section{Experiments}
\textbf{Experimental setup} \; We evaluate a suite of FMs on cloud time series data. We examine several of the most recent and well know FMs for time series: VisionTS \citep{chen2024visionts}, TTM (revision 2) \citep{Ekambaram2024Tiny}, TimesFM \citep{Das2024TimesFM}, Chronos (tiny) \citep{Ansari2024Chronos}, Moirai (small) \citep{Woo2024moirai} and Mamba4Cast \citep{bhethanabhotlamamba4cast}. We look at their performance on datasets constructed out of data released on Huawei Cloud function requests,\footnote{Available at \href{https://github.com/sir-lab/time-series-fm-dataset}{https://github.com/sir-lab/time-series-fm-dataset}} 
as described in Appendix~\ref{appen:dataset}. We name the datasets D1, D2, D3 and D4, each of which correspond to data from a unique data centre in the Huawei Cloud. We note that, none of the FMs looked at have been pretrained on this data or cloud demand data more generally. As explained in Section~\ref{sec:preliminaries}, we perform experiments using a rolling window evaluation. This is done with a context length of $520$ and for three different forecast horizons: $30, 96, 336$. We selected these values as they have been widely used in previous works \citep{Ekambaram2024Tiny, Ansari2024Chronos}.      

\textbf{Baselines} \; We compare the FMs against two simple baselines: \textbf{a)} A per-channel online \textit{linear model} which is fit using ridge regression \citep{hamilton1994Time}. This model is refit on all available data every $200$ time steps. \textbf{b)} A \textit{naive seasonal forecaster} which forecasts by copying repeatedly the last seasonal period in the context as its forecast \citep{Ansari2024Chronos}. For example, for hourly data with a daily periodicity, the naive seasonal forecaster would predict by taking the last 24 hours and repeating it over the forecast window. We note that the naive seasonal forecaster does not need to have access to any training data. More details on the baselines can be found in Appendix~\ref{appen:baselines}. 

\textbf{Metrics} \; We report the MASE of each approach, a common time series evaluation metric \citep{Ansari2024Chronos}. MASE is the mean absolute error (MAE) normalized by dividing by the MAE of a naive seasonal forecaster over the context, as detailed in Appendix~\ref{appen:metrics}. This normalisation adjusts for varying scales across channels and over time. \textit{We also report results using RMSSE in Appendix~\ref{appen:rmsse}}, which is defined analogously to MASE but with RMSE instead of MAE \citep{hyndman2006another}. Both metrics yield consistent conclusions.

\begin{table*}[t]
\setlength\tabcolsep{2.5pt}
\centering
\caption{\textbf{MASE results of zero-shot FMs forecasts for cloud data compared to the baselines of an online linear model and a naive seasonal forecaster.} The table shows that the baseline methods perform the best across all datasets and forecast horizons ($H$), demonstrating that currently, zero-shot FMs struggle to perform well on cloud data.}
\label{table:main_results}
\begin{tabular}{@{}cc|cccccccc@{}}
\toprule
\addlinespace
& \multicolumn{1}{c}{} & & & \multicolumn{6}{c}{Zero-Shot FMs} \\
\cmidrule(l){5-10}
\multirow{2}{*}{Dataset} &  \multirow{2}{*}{$H$}  & \multirow{2}{*}{\textbf{Linear}} & \multirow{2}{*}{\textbf{Seasonal}} & \multirow{2}{*}{\textbf{VisionTS}} & \multirow{2}{*}{\textbf{TTM}} & \multirow{2}{*}{\textbf{TimesFM}} & \multirow{2}{*}{\textbf{Chronos}} & \multirow{2}{*}{\textbf{Moirai}} & \multirow{2}{*}{\textbf{Mamba4Cast}} \\ 
& & \\
\midrule
\multirow{3}{*}{D1} & 30 & \textbf{1.404} & 1.577 & 1.756 & 1.596 & 3.020 & 2.457 & 1.972 & 2.837  \\
& 96 & 1.994 & \textbf{1.875} & 1.966 & 2.275 & 5.754 & 5.622 & 2.656 & 5.618  \\
& 336 & 3.175& \textbf{2.809} & 2.872 & 3.398 & 7.225 & 8.068 & 3.585 & 7.565  \\
\addlinespace
\multirow{3}{*}{D2} & 30 & \textbf{1.170} & 1.231 & 1.686 & 1.844 & 2.772 & 2.454 & 2.443 & 3.207 \\
& 96 & \textbf{1.550} & 1.398 & 1.711 & 2.396 & 5.057 & 4.342 & 3.642 & 6.718 \\
& 336 & 2.262 & \textbf{1.818} & 2.107 & 2.999 & 6.346 & 5.725 & 3.983 & 13.268  \\
\addlinespace
\multirow{3}{*}{D3} & 30 & \textbf{1.082} & 1.182 & 1.510 & 1.747 & 2.181 & 2.037 & 2.187 & 2.500 \\
& 96 & \textbf{1.153} & 1.179 & 1.431 & 1.943 & 3.466 & 2.806 & 2.682 & 3.994 \\
& 336 & 1.290 & \textbf{1.236} & 1.452 & 2.128 & 3.909 & 3.260 & 2.573 & 5.699 \\
\addlinespace
\multirow{3}{*}{D4} & 30 & 1.578 & \textbf{1.349} & 1.618 & 2.618 & 2.731 & 2.951 & 2.843 & 3.546  \\
& 96 & 1.578 & \textbf{1.299} & 1.424 & 2.369 & 2.667 & 2.852 & 2.605 & 3.917 \\
& 336 & 1.695 & \textbf{1.307} & 1.414 & 2.368 & 2.657 & 2.802 & 2.599 & 5.228  \\
\addlinespace
\bottomrule
\end{tabular}
\end{table*}

\textbf{Results} \; The main results of our experiments are displayed in Table~\ref{table:main_results}. The table shows that none of the FMs being evaluated perform as well as either of the simple baselines we compare to. For example, the naive seasonal forecaster performs better than all the FMs across all datasets and forecast horizons. Moreover, the performance difference is often large; for example, the naive seasonal forecaster incurs a MASE typically half that of TimesFM. This suggests that currently zero-shot FMs do not generalise well to cloud data. We also note that alongside the good predictive performance of the baseline methods they are also the most computationally efficient, as demonstrated in Appendix~\ref{appen:compCost}.
\begin{figure}[t]
    \centering
    \begin{minipage}[b]{0.49\textwidth}
        \centering
        \includegraphics[width=\textwidth]{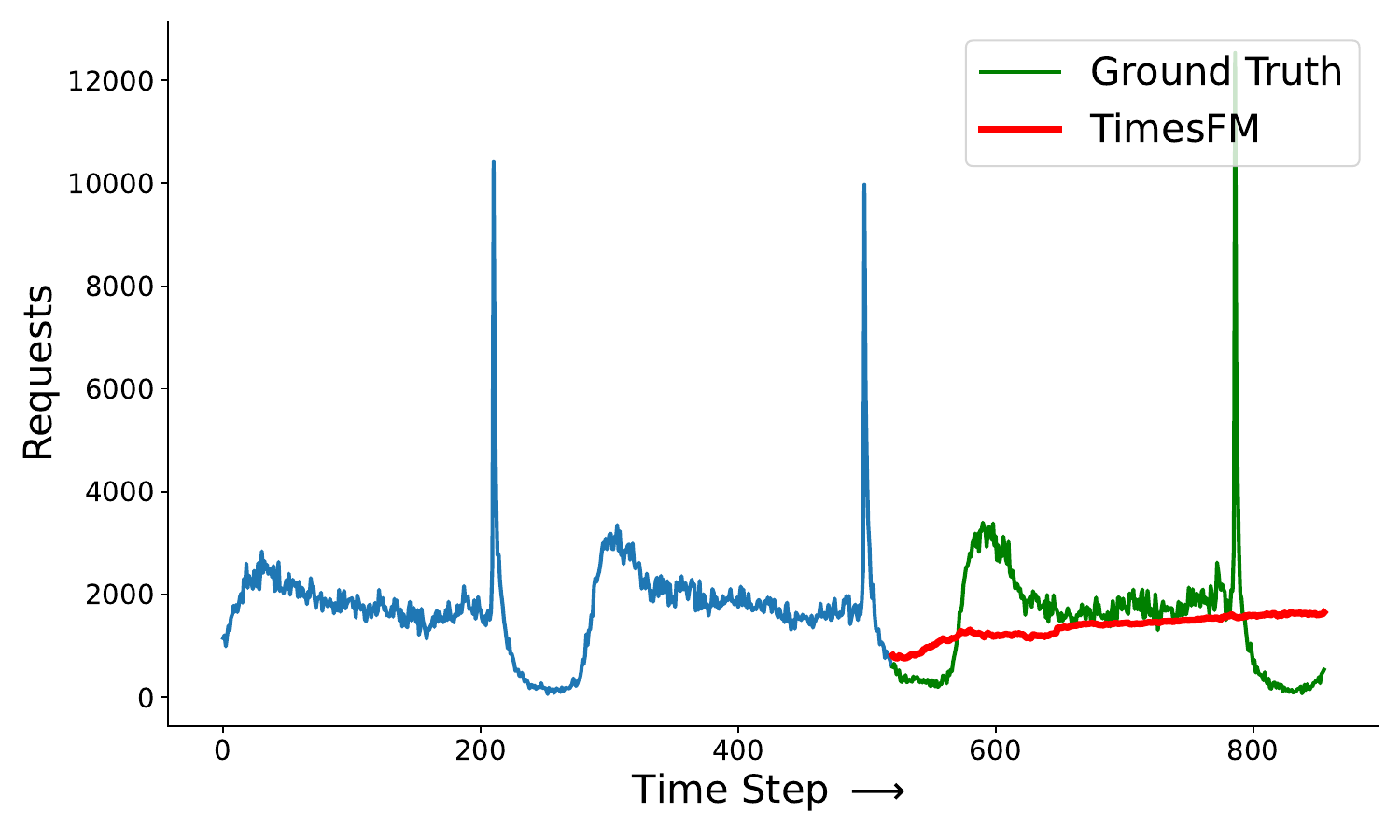}
    \end{minipage}
    \hfill
    \begin{minipage}[b]{0.49\textwidth}
        \centering
        \includegraphics[width=\textwidth]{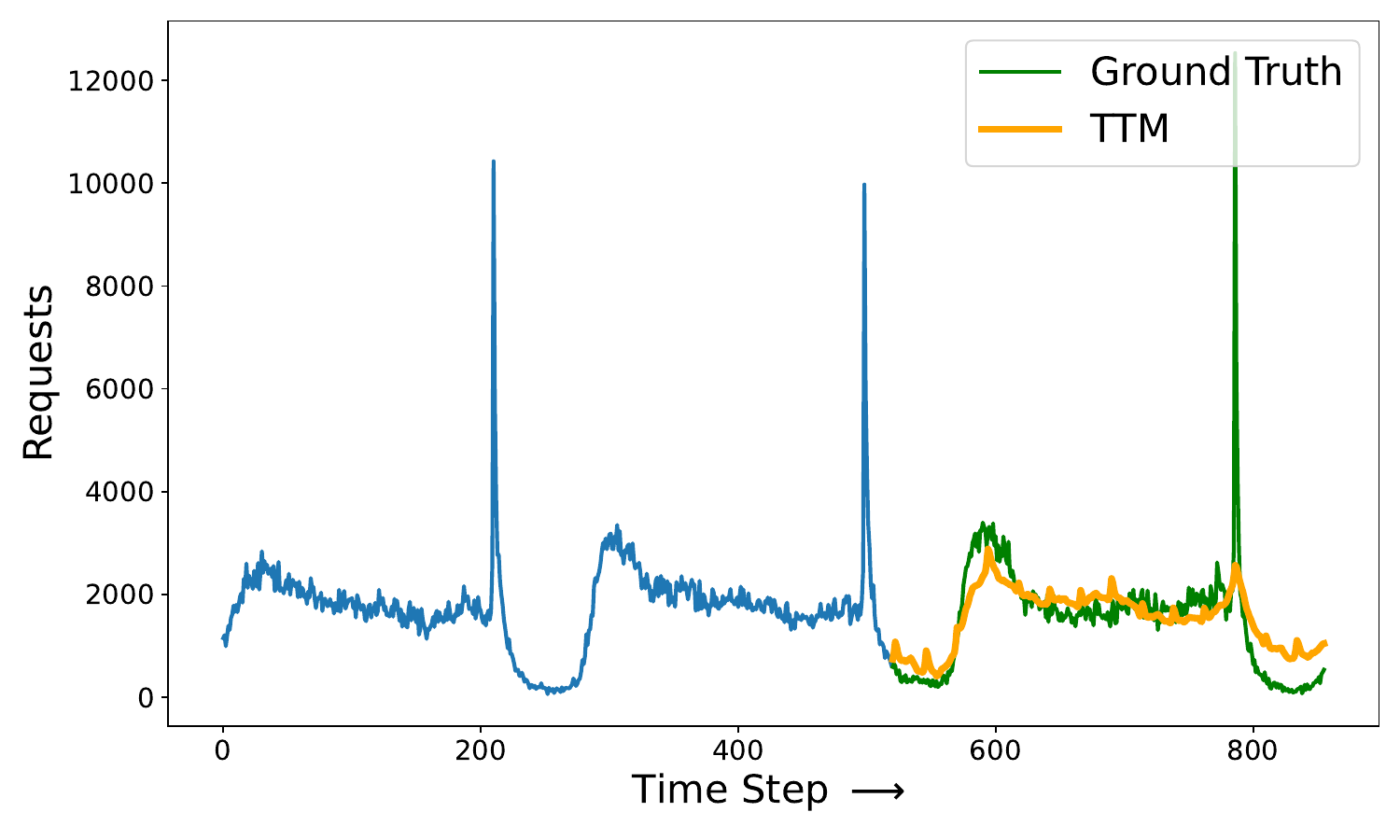}
    \end{minipage}
    \vskip\baselineskip
    \begin{minipage}[b]{0.49\textwidth}
        \centering
        \includegraphics[width=\textwidth]{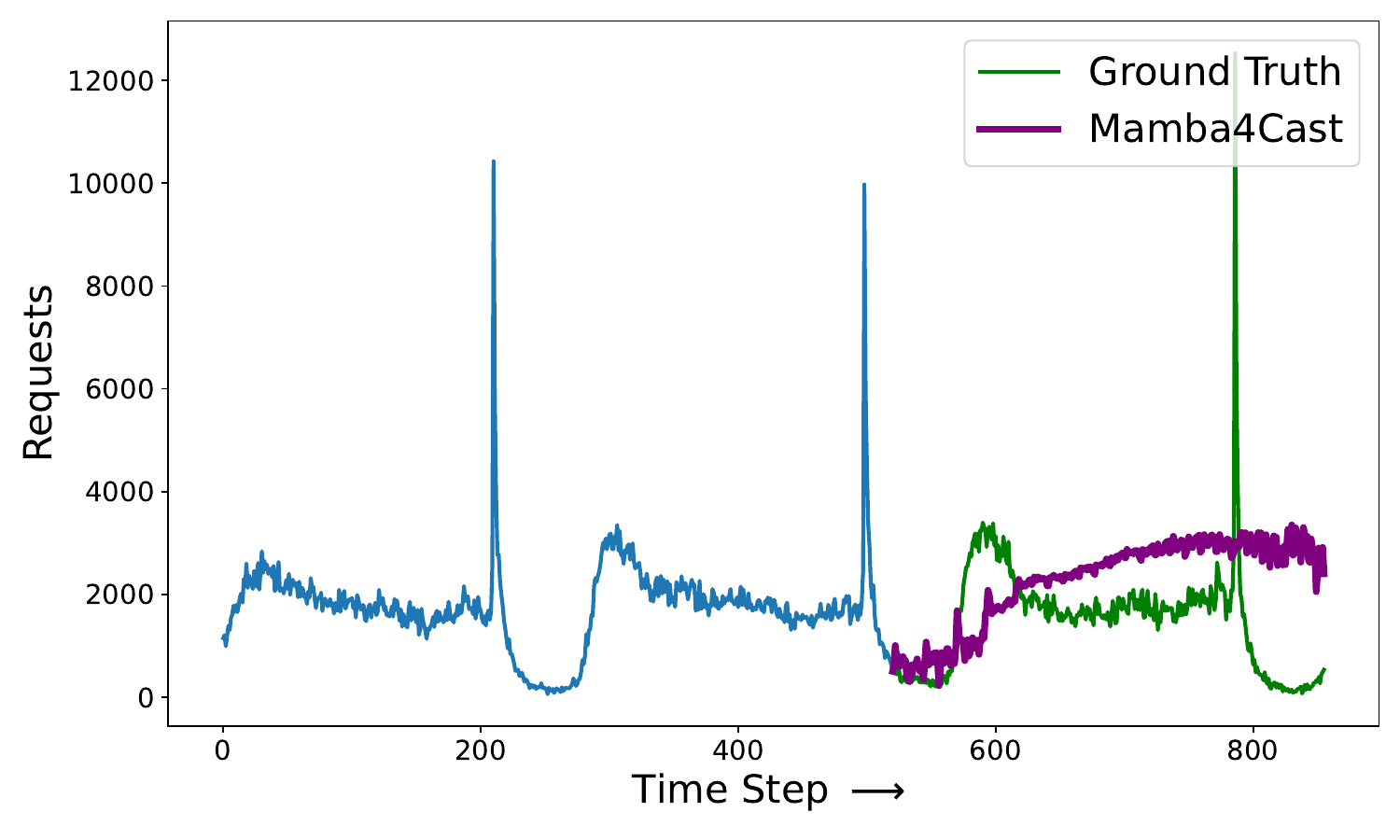}
    \end{minipage}
    \hfill
    \begin{minipage}[b]{0.49\textwidth}
        \centering
        \includegraphics[width=\textwidth]{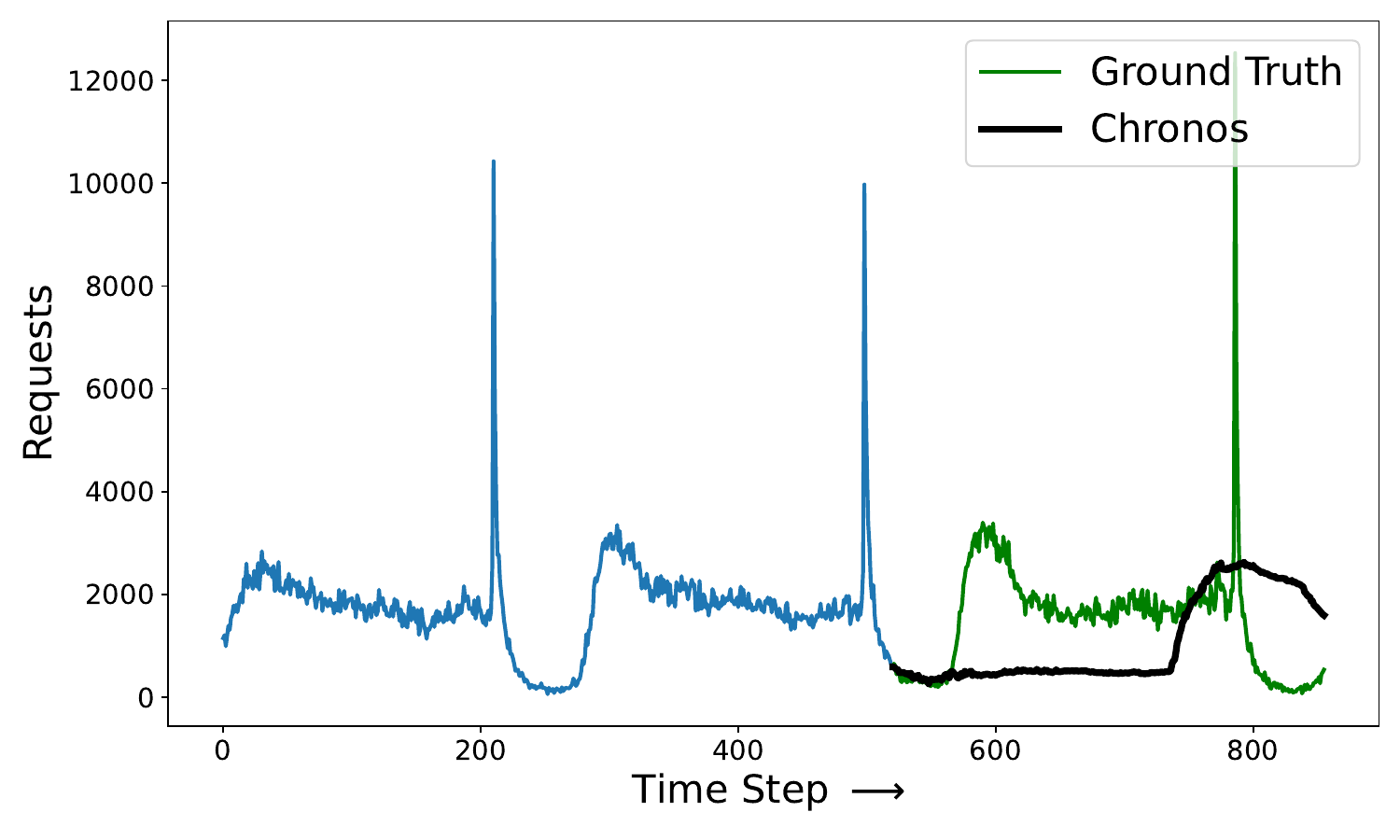}
    \end{minipage}
    \caption{\textbf{FM forecasts at the same time-step ($\bm{t=6030}$) for the third channel of the Huawei Cloud D2 dataset.} The plots shows the forecasts for TimesFM, TTM, Mamba4Cast and Chronos FMs. None of the FMs forecasts are accurate. TimesFM, Mamba4Cast and Chronos do not identify the seasonal pattern. While, TTM performs better than the rest but does not forecasts the periodic spike in demand, incurring a large inaccuracy at that point.}
    \label{fig:FM_forecasts}
\end{figure}

\subsection{What do Zero-Shot FM Forecasts Look Like?} \label{sec:why}
To analyse further the poor performance of zero-shot FMs in the cloud domain we plot a typical failure case of FMs in Figure~\ref{fig:FM_forecasts}. The figure displays the forecasts for TimesFM, TTM, Mamaba4Cast and Chronos on the third channel of the D2 dataset---forecasts for Moirai and VisionTS are given in Appendix~\ref{appen:add_forecasts}. The context (shown by the blue curve) is standard of cloud time series, having a strong seasonal pattern and large spikes \citep{diao2024forecasting}, and the ground truth is roughly a repeat of the seasonal pattern. However, despite the regularity of the time series, all of the FMs give poor predictions, appearing to not make use of the clearly periodic behaviour. For example, three of the FMs, in Figure~\ref{fig:FM_forecasts}, do not identify the seasonal pattern at all; while the other, TTM, gets the rough shape correct but misses the spikes. We note that, in cloud provision settings predicting demand spikes is vital, a failure to do so can degrade service quality.

Overall, the example forecasts in Figures~\ref{fig:moirai} and \ref{fig:FM_forecasts} show that for typical cloud data, zero-shot FM forecasts can fail to resemble the ground truth, even when the ground truth is readily predictable from the context given---as demonstrated by the baseline forecasts given in Appendix~\ref{appen:add_forecasts}. We emphasise that the failures shown are not unique to the displayed channels or time steps, being commonplace in our experiments, evidencing the poor overall performance of FMs on cloud data shown in Table~\ref{table:main_results}.
\begin{wrapfigure}{r}{0.5\textwidth}
    \centering
    \includegraphics[width=0.49\columnwidth]{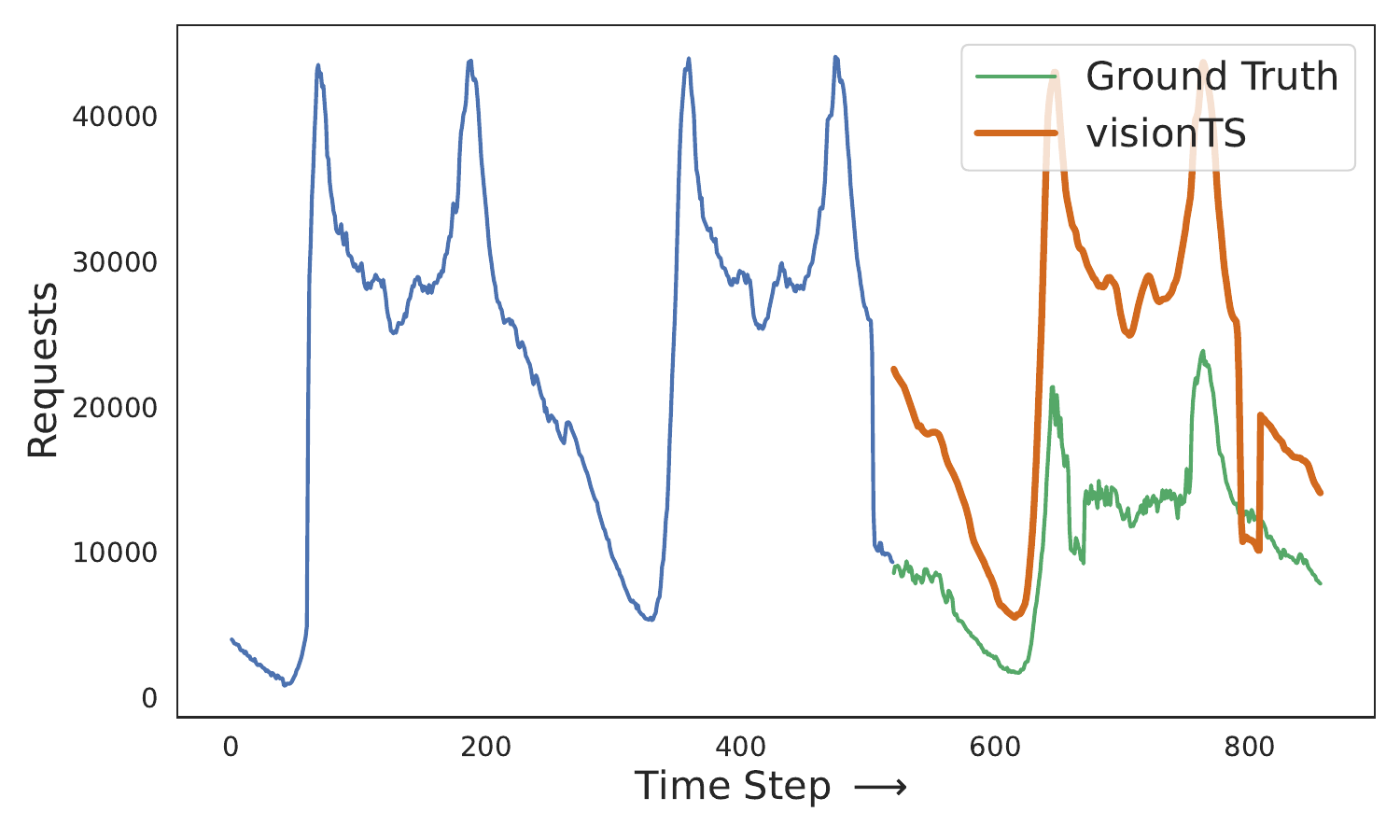}
    \caption{\textbf{An example forecast for VisionTS, demonstrating that it is roughly equivalent to a naive seasonal forecaster for cloud data.} Because the VisionTS forecast is roughly the same as the last seasonal period in the context, it is very similar to the naive seasonal forecast.}
    \label{fig:visionts}
    \vspace{-22mm}
\end{wrapfigure}

\textbf{VisionTS} \; often performs quite well relative to the other FMs in our experiments (see Table~\ref{table:main_results}). While its performance is still worse than the baselines, it is worth exploring why this is. We find in our explorations that VisionTS behaves very similarly to a naive seasonal forecaster in most cases, simply coping the context as its forecast. An example of this behaviour is shown in Figure~\ref{fig:visionts}. Since most channels are relatively periodic this strategy performs well, largely explaining the good performance of this model. To analyse this further, we computed the normalised (zero-lag) cross-correlation between the forecasts of VisionTS and the naive seasonal forecaster on D1 \textit{obtaining a median \\ correlation of 0.9992}. 

\section{Conclusions}
In this work we examined the performance of zero-shot time series foundation models (FMs) in the cloud domain. While time series FMs claim to generalise to many domains including cloud \citep{Woo2024moirai}, we find that FMs perform worse than the simple baselines of an online linear model and naive seasonal forecaster on cloud data. We also present evidence of pathological behaviours of FMs, demonstrating they can produce chaotic (as in Figure~\ref{fig:moirai}) or illogical (as in Figure~\ref{fig:FM_forecasts}) forecasts. Finally, we show that the best performing FM, VisionTS, performs well due to giving very similar forecasts to a naive seasonal forecaster. The results suggest that the particular characteristics of cloud data---such as spikiness \citep{diao2024forecasting}---can make FMs produce bad zero-shot forecasts. This questions their current applicability to cloud forecasting problems.

\subsubsection*{Acknowledgments}
We would like to kindly thank Luke Darlow for his help and Thomas~L. Lee would like to thank his PhD supervisor, Amos Storkey, for his general guidance on research practice. 

\FloatBarrier
\bibliography{references}
\bibliographystyle{iclr2025}

\FloatBarrier
\newpage
\appendix
\section{Generalization in Zero-Shot Time Series Forecasting}
While in the main text we have aimed to show if and how zero-shot FMs fail to forecast cloud data here we examine the question of why. To do this we ask more generally how much can we hope to generalize in time series forecasting? In both the vision and text domain it has been shown that zero-shot FMs have good generalization capabilities \citep{minaee2024large, zhou2022domain} and hence it can be natural to suggest by using zero-shot FMs in the time domain we can achieve similar generalization capabilities. However, there are reasons to question this belief, mainly due to less constrained nature of the continuation of a time series versus the continuation of an image (i.e. inpainting) or text generation. For instance, given the same context there maybe many valid forecasts based on the time series domain the data was generated from. More concretely, in cloud data there are spikes which happen at regular intervals which should be forecasted but in other domains spikes could be caused by noise and so should not appear in the forecast. This suggests that as the claims of zero-shot time series FM generalization expands it is quickly apparent that there is no free lunch. Hence, claims of widespread generalization ability or generalization to untested domains for zero-shot forecasts should be inspected carefully. 

Additionally, the above suggest that to create well performing FMs for time series data there needs to be a mechanism for \emph{conditioning} on previous data from the time series being forecasted or from similar time series. This can be straightforwardly achieved by finetuning methods. However, this can be compute and time intensive and requires access to a significant amount of offline data. So, in real-world scenarios where time series data is generated online, shifts in distribution and compute/training time is constrained, there is a need to look at new conditioning mechanisms. This has already been looked at in some works \citep{lee2025lightweight, Ma2023} but there is scope to do more in this area which could realize the goal of general time series FMs.    

\section{Additional Experimental Details}
\subsection{Construction of the Huawei Cloud Datasets} \label{appen:dataset}
\begin{table}[h]
\centering
\caption{\textbf{Dataset statistics and the seasonalities used for each dataset in the computation of MASE.}}
\label{tab:DatasetStats}
\begin{tabular}{c|ccccp{25mm}}
\toprule
\multirow{2}{*}{\textbf{Dataset}} & \multirow{2}{*}{\textbf{Time-Step Period}} & \multirow{2}{*}{\textbf{Seasonality}} & \multirow{2}{*}{\textbf{\#Channels}} & \multirow{2}{*}{\textbf{\#Time Steps}} & \multirow{2}{*}{\textbf{Selected channels}} \\
\\ 
\midrule
\addlinespace
D1 & 5 mins & 288 & 17 & 8640 & $[365, 917, 919,\newline 920, 921, 926,\newline 929, 1016, 1289,\newline 1235, 976, 504,\newline 745, 889, 923,\newline 958, 961]$\\
\addlinespace
D2 & 5 mins & 288 & 18 & 8640 & $[965, 1089, 1175,\newline 945, 450, 1017,\newline 1346, 1017, 1346,\newline 1132, 932, 909,\newline 1478, 1465, 949,\newline 1499, 975, 1213]$ \\
\addlinespace
D3 & 5 mins & 288 & 7 & 8640 & $[2310, 2214, 2223,\newline 129, 2472, 2461,\newline 2471]$ \\
\addlinespace
D4 & 5 mins & 288 & 8 & 8640 & $[449, 444, 435,\newline 487, 447, 438,\newline 455, 426]$\\
\addlinespace
\bottomrule
\end{tabular}
\end{table}
To look at how well zero-shot FMs perform on cloud time series we construct a cloud dataset out of publicly accessible data from the Huawei Cloud.\footnote{The full Huawei Cloud data release is available at \href{https://github.com/sir-lab/data-release}{https://github.com/sir-lab/data-release}} This data consists of function requests from Huawei's serverless cloud system. The data is divided into the function requests across the data centres of five regions in China. As described in more detail in \citet{joosen2024serverless}, the data is typical of the cloud domain having large spikes and strong periodicity \citep{diao2024forecasting, darlow2024dam, aksu2024gift}. From this large store of data we curate four manageable datasets from the first two and last two regions, respectively, calling them D1, D2, D3 and D4. We disregard the third region which is much smaller than the other region's datasets\footnote{E.g., region one has $2487$ functions while Region three has $237$ functions} and is generally much sparser. From the regions we selected, we downsample the number of channels by selecting ones which have a large amount of activity. We ignore channels which are near-identical copies of other selected channels, have very low activity and/or are highly random, making them unsuitable for forecasting. The indices of the selected channels are shown in the \emph{selected channels} column of Table~\ref{tab:DatasetStats}. After channel selection, the data is cleaned by filling missing values with zeros. It is then downsampled to 5-minute intervals by summing the values within each interval. Importantly, the selection of the channels and other dataset construction decisions was finalised \textit{before} studying any forecast models and the dataset was not subsequently altered in any way when studying the performance of the FMs. We have released this dataset at \href{https://github.com/sir-lab/time-series-fm-dataset}{https://github.com/sir-lab/time-series-fm-dataset} and 
an overview of the statistics of each dataset is shown in Table~\ref{tab:DatasetStats}.

\subsection{Details of Baselines} \label{appen:baselines}
The baselines we employ are simple, and we have included the main details in the text. However, there are three additional implementation details to mention: \textbf{a)} For numerical stability, we standardise the data when fitting the online linear model. This is achieved by normalising the data with a running standard deviation for each channel. The running standard deviation is updated online using Welford's algorithm \citep{welford1962note}. At inference time no data standardisation is required---as $\alpha W\left(\frac{\bm{x}}{\alpha}\right) = W\bm{x}$ for any $\alpha$ and where $W$ is a matrix \citep{toneranalysis}. \textbf{b)} The ridge regression regularisation coefficient used when fitting the online linear model is set to $0.5$ for all experiments. \textbf{c)} Before the linear model is fit for the first time it will give forecasts based on how it was initialised, in this work we initialise it to be the naive seasonal forecaster.

\subsection{Details of FMs} 
We have aimed to used the same hyperparameters/settings as the original works for each FM looked at. However, both TTM and TimesFM were trained for a context length of $512$ and therefore when using them we drop the first $8$ values of each $520$-long context. Also, for TTM the currently released models only predict to a horizon length of $96$. Therefore, to generate the forecasts for the horizon length of $336$ we auto-regressively feed-in the constructed forecast, as done in the paper proposing TTM \citep{Ekambaram2024Tiny}.     

\subsection{Metrics}\label{appen:metrics}
The Mean Absolute Scaled Error (MASE) \citep{hyndman2006another} is for a given (per-channel) forecast $\bm{\hat{y}}$, context $\bm{x}$ and ground truth forecast $\bm{y}$ calculated as
\begin{align*}
    \text{MASE}(\bm{\hat{y}}, \bm{y}, \bm{x}) = \frac{L-S}{T} \frac{\sum_{i=1}^{T} |\hat{y}_i - y_i|}{\sum_{i=1}^{H-S}|x_i-x_{i+S}|},
\end{align*}
where $S$ represents the seasonality. As explained in the main text, this the MAE normalised using the MAE of the naive seasonal forecaster computed on the context.  

The Root Mean Squared Scaled Error (RMSSE) metric \citep{hyndman2006another} is computed in the same way as MASE but with MAE replaced with MSE and with the whole formula rooted:
\begin{align*}
    \text{RMSSE}(\bm{\hat{y}}, \bm{y}, \bm{x}) = \sqrt{\frac{L-S}{T} \frac{\sum_{i=1}^{T} (\hat{y}_i - y_i)^2}{\sum_{i=1}^{H-S}(x_i-x_{i+S})^2}}.
\end{align*}

\newpage
\section{Additional Experiments}
\subsection{Computational Cost}\label{appen:compCost}
\begin{table*}[h]
\centering
\caption{\textbf{Run-time of the online linear model and a naive seasonal forecaster baselines compared to the most efficient zero-shot FM, TTM.} We perform the experiments on two Intel(R) Xeon(R) Platinum 8168 CPUs. The table shows that alongside the baselines being the most accurate forecasters, they are also the most computationally efficient.}
\label{table:comp_cost}
\begin{tabular}{@{}cc|ccc@{}}
\toprule
\addlinespace
& \multicolumn{1}{c}{} & \multicolumn{3}{c}{Run-Time (seconds)} \\
\cmidrule(l){3-5}
\multirow{2}{*}{Dataset} &  \multirow{2}{*}{$H$}  & \multirow{2}{*}{\textbf{Linear}} &  \multirow{2}{*}{\textbf{Seasonal}} & \multirow{2}{*}{\textbf{TTM}} \\ 
& & \\
\midrule
\multirow{3}{*}{D1} & 30 & 30 & 20 & 52  \\
& 96 & 33 & 22 & 53 \\
& 336 & 39 & 28 & 202 \\
\addlinespace
\multirow{3}{*}{D2} & 30 & 32 & 22 & 50 \\
& 96 & 34 & 23 & 44 \\
& 336 & 41 & 29 & 213 \\
\addlinespace
\multirow{3}{*}{D3} & 30 &  12 & 8 &  17 \\
& 96 & 13 & 9 & 17  \\
& 336 & 16 & 11 & 73  \\
\addlinespace
\multirow{3}{*}{D4} & 30 & 15 & 10 & 25  \\
& 96 & 16 & 10 & 18 \\
& 336 & 18 & 13 & 65  \\
\addlinespace
\bottomrule
\end{tabular}
\end{table*}

\newpage
\FloatBarrier
\subsection{Baseline and Other FM Forecasts for Figure~\ref{fig:FM_forecasts}} \label{appen:add_forecasts}
\begin{figure}[h]
    \centering
    \begin{minipage}[b]{0.49\textwidth}
        \centering
        \includegraphics[width=\textwidth]{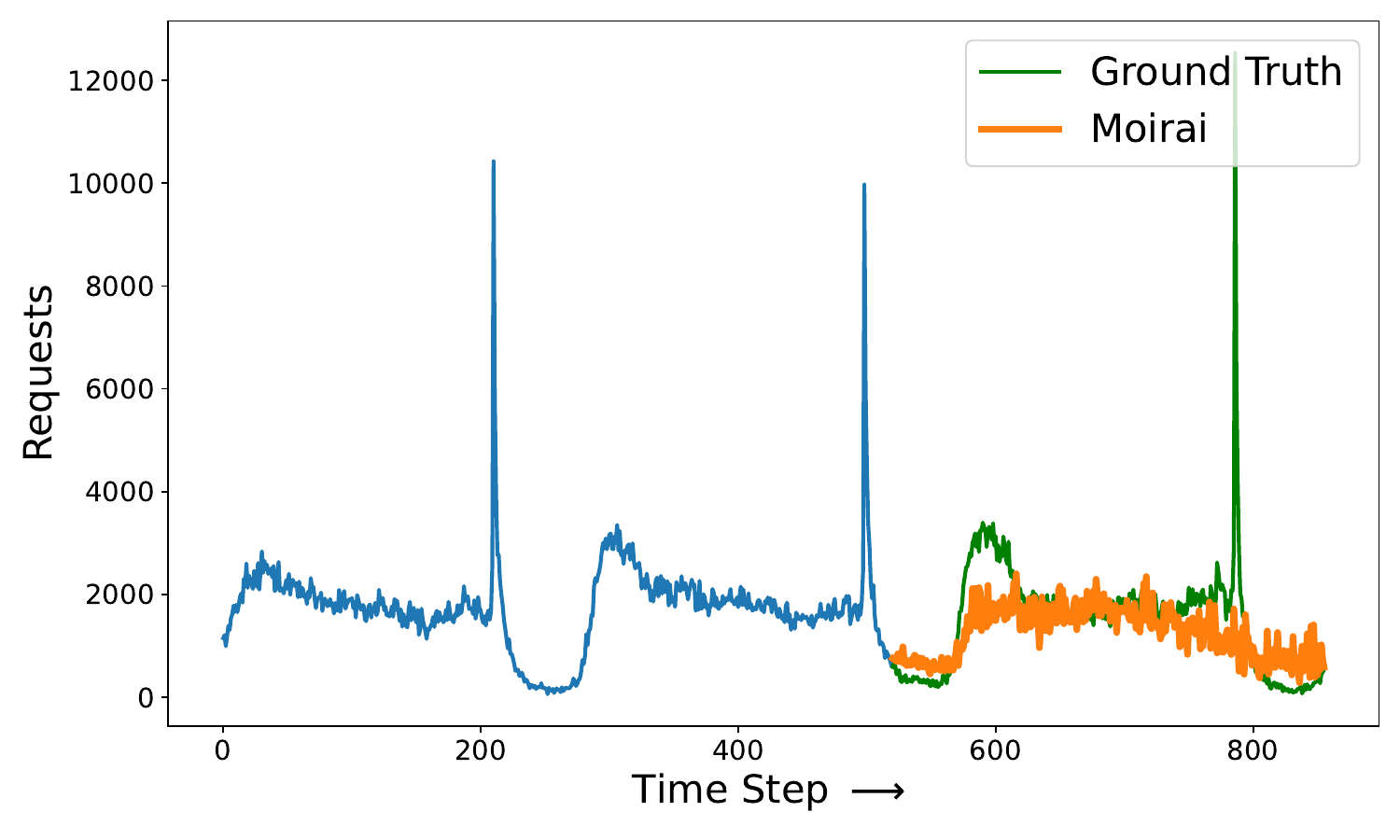}
    \end{minipage}
    \hfill
    \begin{minipage}[b]{0.49\textwidth}
        \centering
        \includegraphics[width=\textwidth]{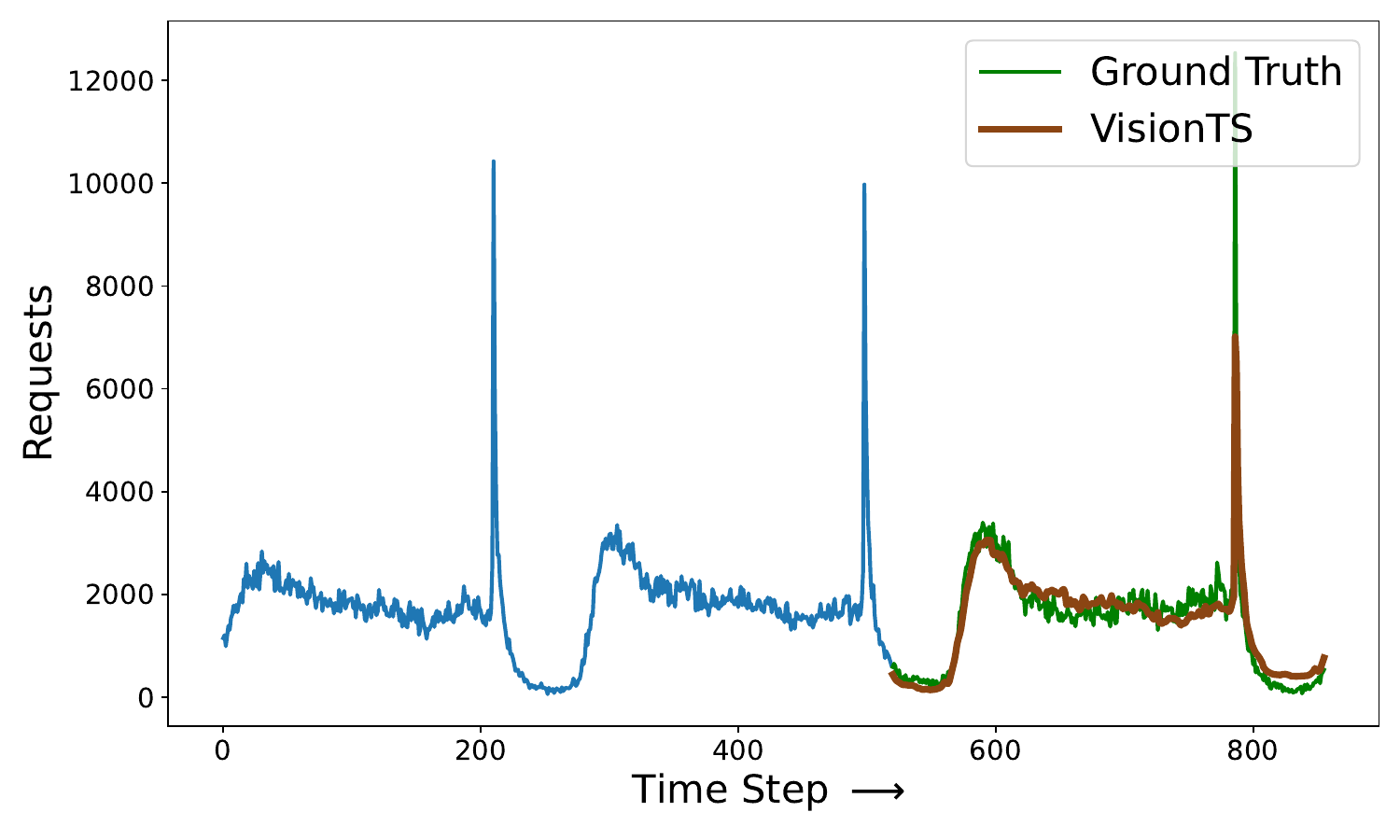}
    \end{minipage}
    \begin{minipage}[b]{0.49\textwidth}
        \centering
        \includegraphics[width=\textwidth]{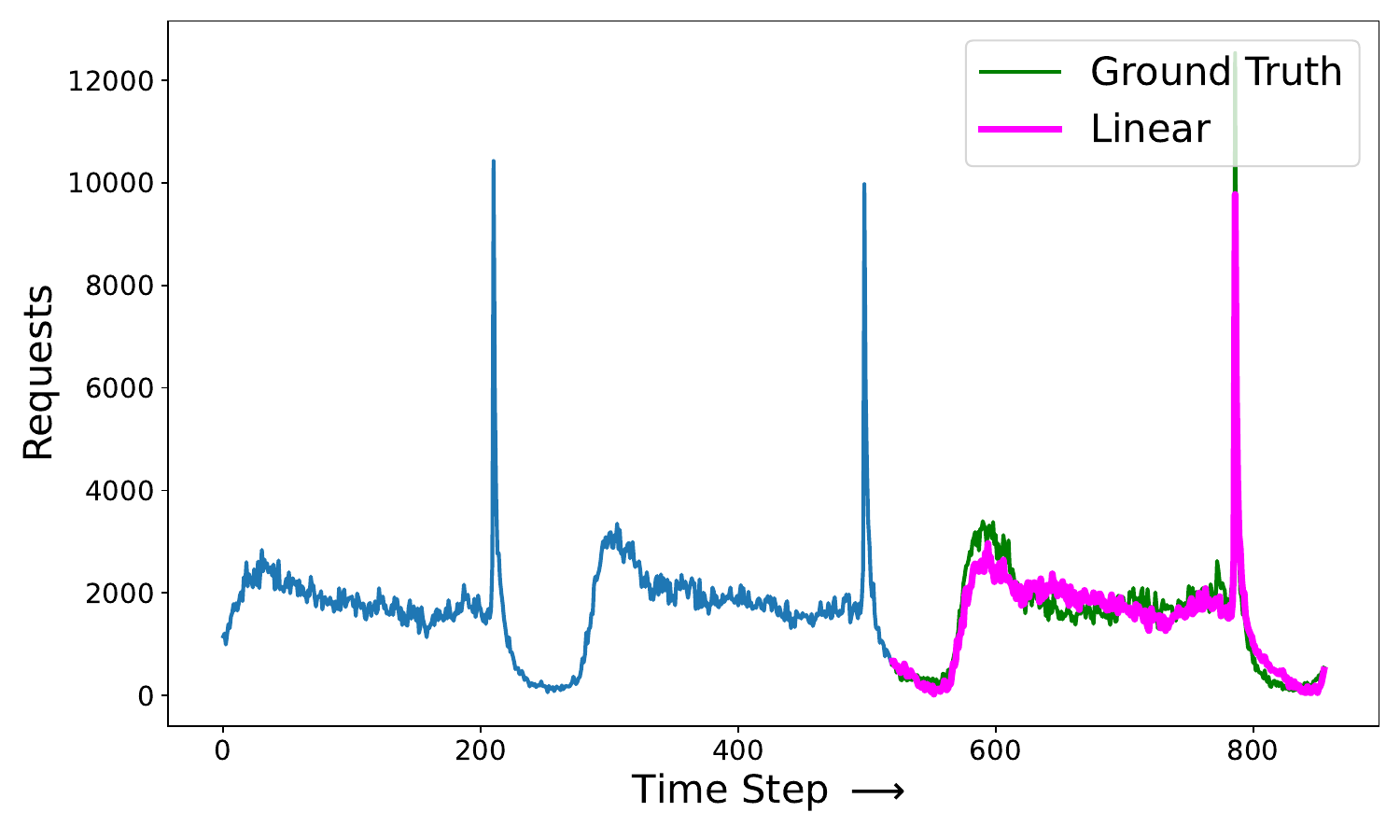}
    \end{minipage}
    \hfill
    \begin{minipage}[b]{0.49\textwidth}
        \centering
        \includegraphics[width=\textwidth]{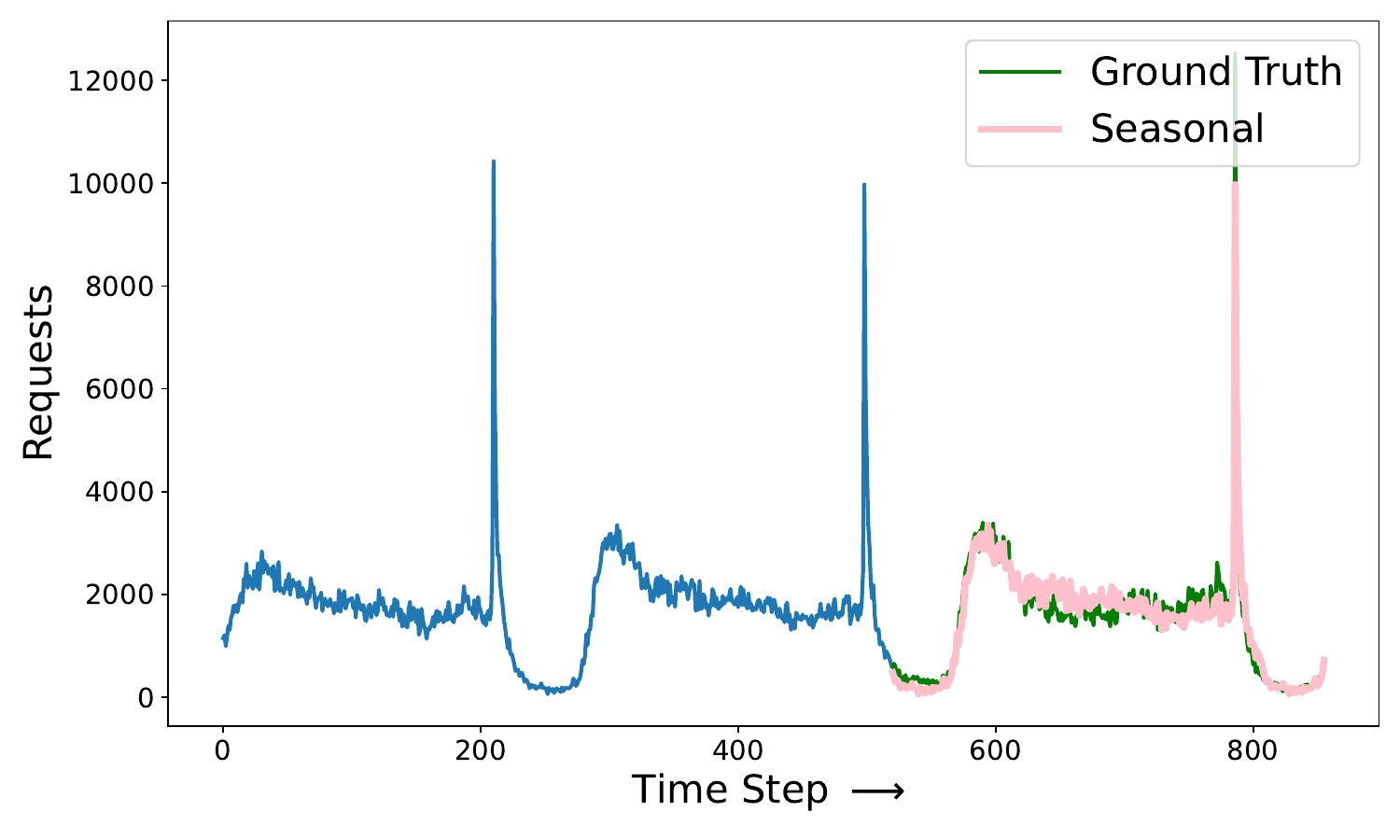}
    \end{minipage}
    \caption{\textbf{FM and baseline forecasts at the same time-step ($\bm{t=6030}$) for the third channel of the Huawei Cloud D2 dataset.} These plots show the forecasts for FMs not shown in Figure~\ref{fig:FM_forecasts}: Moirai and VisionTS. It also shows the forecasts for the two baseline models: the online linear and naive seasonal forecasters. As in Figure~\ref{fig:FM_forecasts} the FMs do not perform as well as the baselines. Moirai gives a poor forecast and does not predict the spike. While, as discussed in Section\ref{sec:why}, VisionTS gives a similar---but dampened and so worse---forecast to the naive seasonal forecaster. This is in contrast to the two baseline methods which predict accurately, importantly forecasting the spike in the data.}
    \label{fig:add_forecasts}
\end{figure}

\newpage
\FloatBarrier
\subsection{Additional Forecasts}
\begin{figure}[h]
    \centering
    \begin{minipage}[b]{0.49\textwidth}
        \centering
        \includegraphics[width=\textwidth]{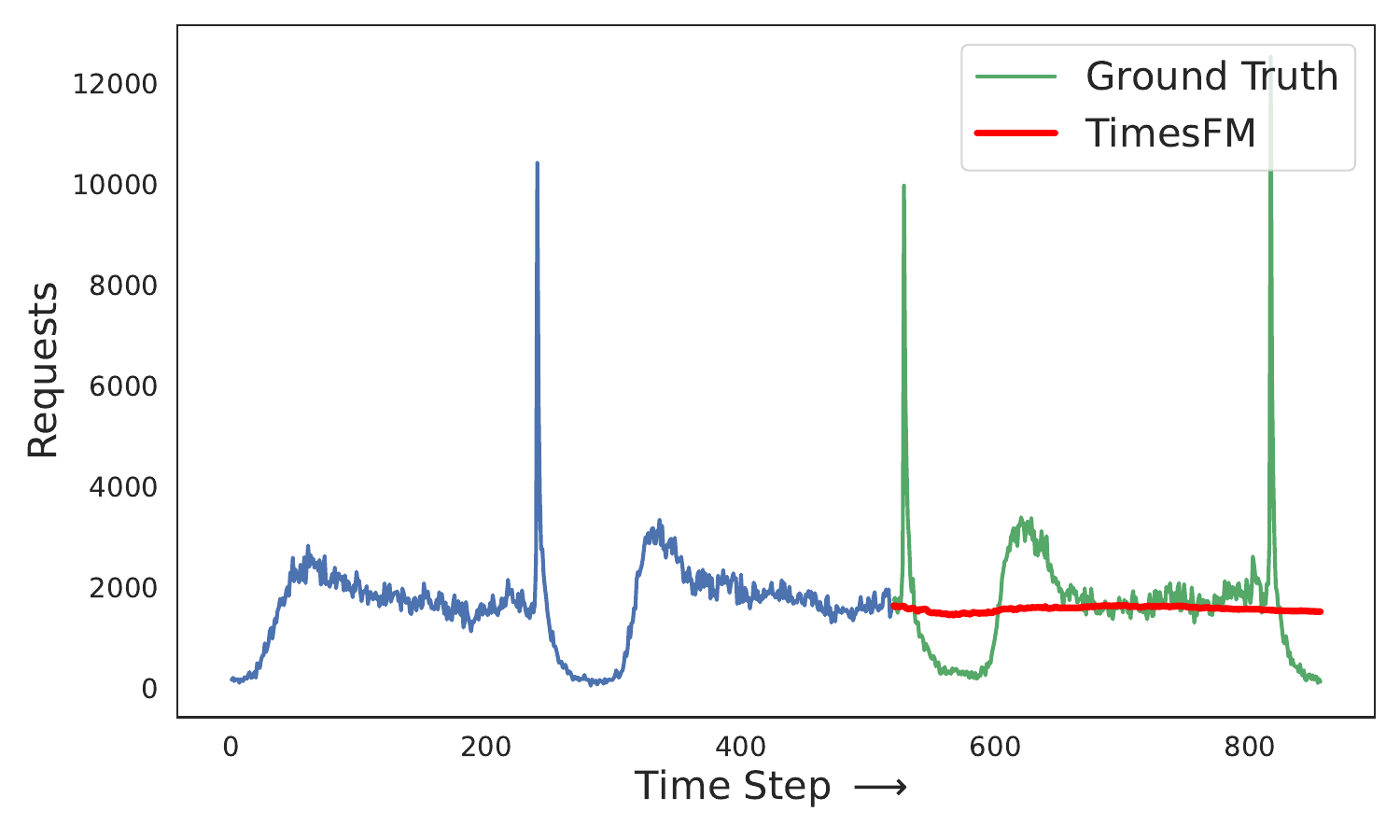}
    \end{minipage}
    \hfill
    \begin{minipage}[b]{0.49\textwidth}
        \centering
        \includegraphics[width=\textwidth]{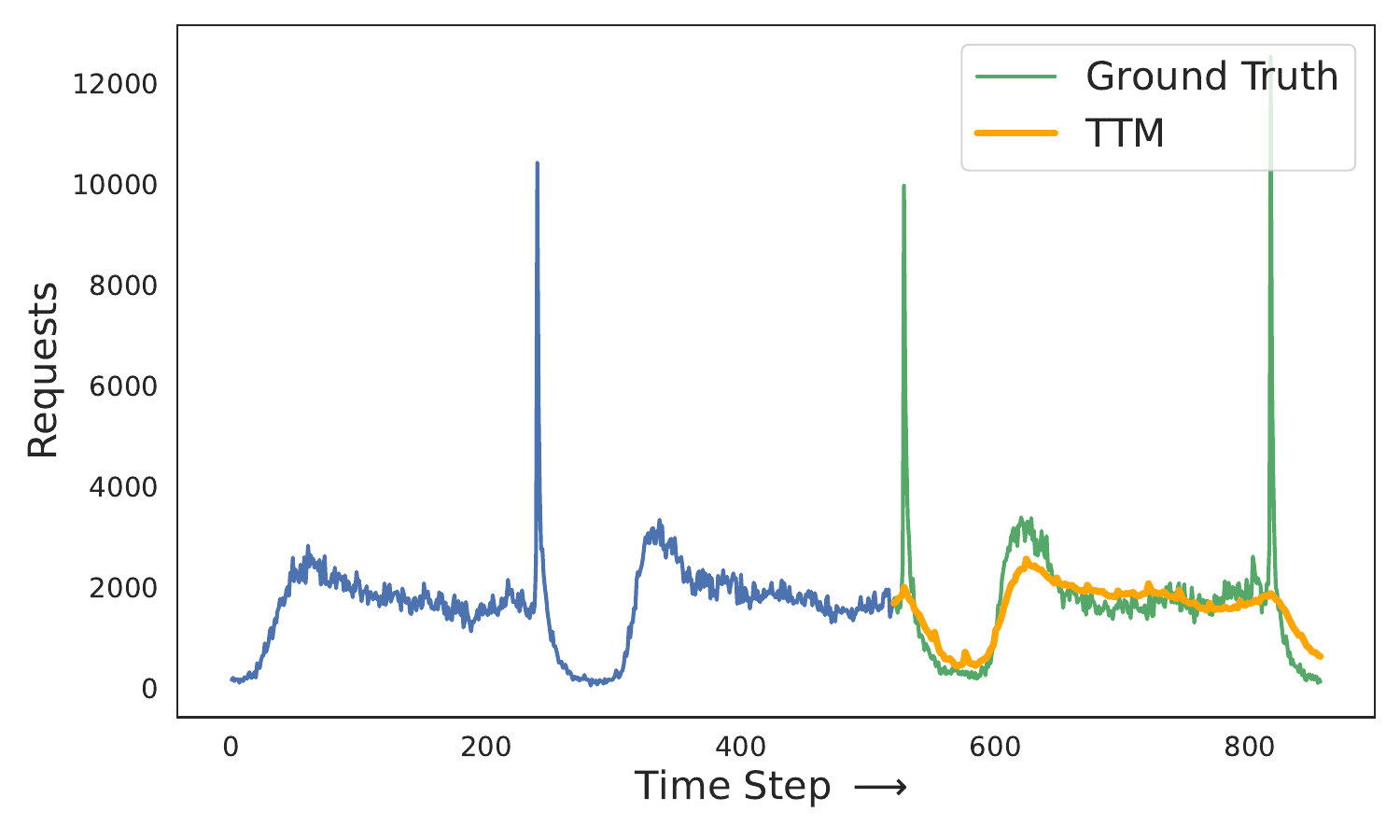}
    \end{minipage}
    \begin{minipage}[b]{0.49\textwidth}
        \centering
        \includegraphics[width=\textwidth]{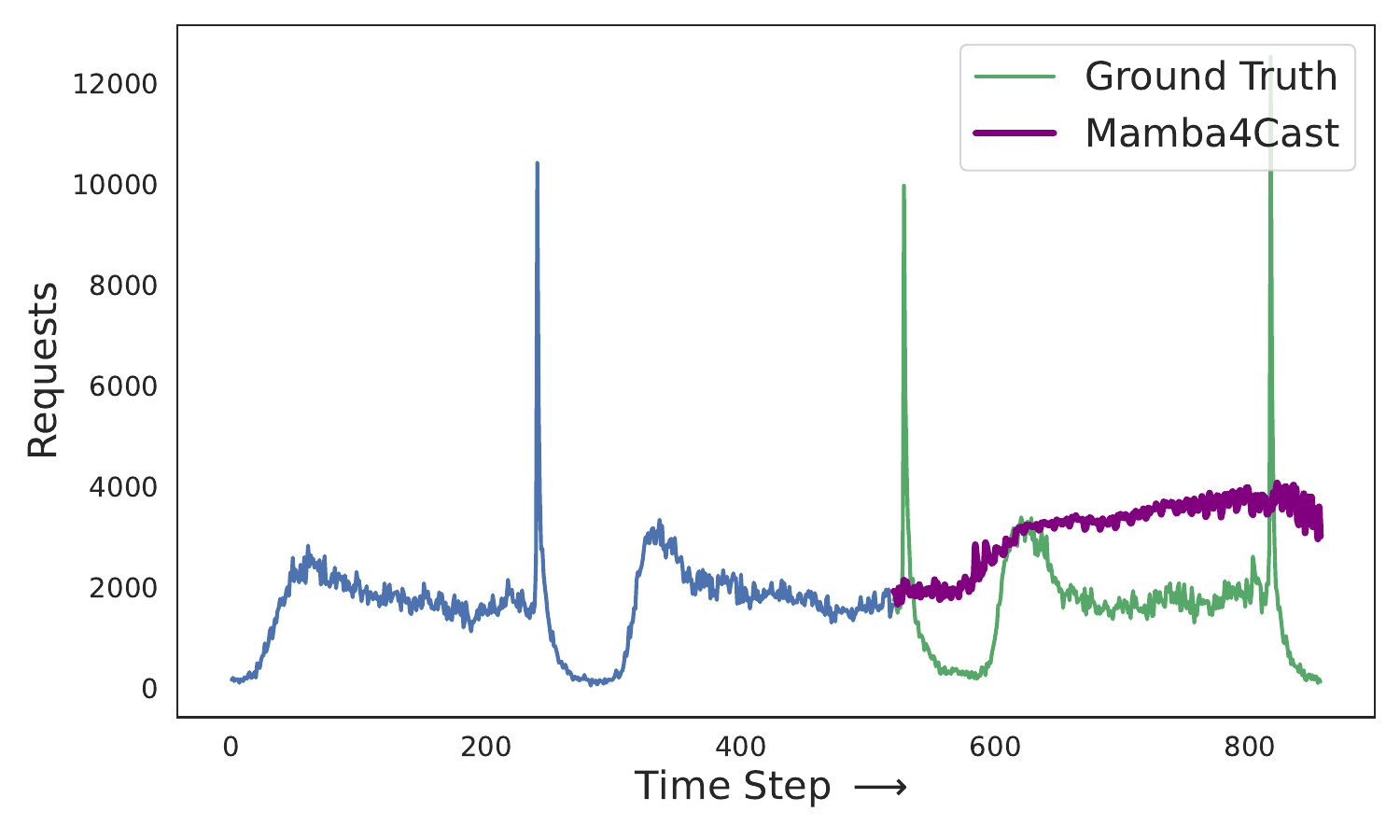}
    \end{minipage}
    \hfill
    \begin{minipage}[b]{0.49\textwidth}
        \centering
        \includegraphics[width=\textwidth]{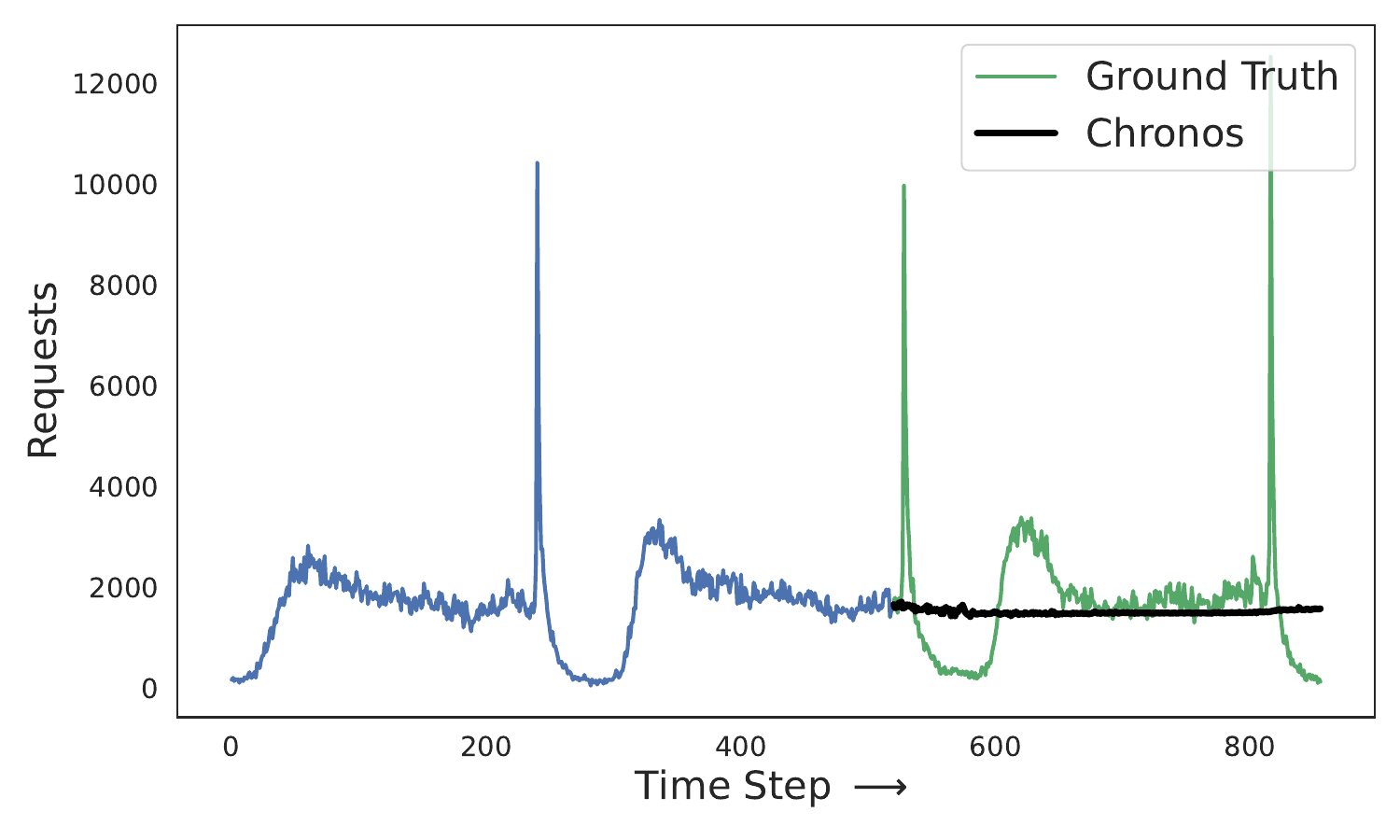}
    \end{minipage}
    \begin{minipage}[b]{0.49\textwidth}
        \centering
        \includegraphics[width=\textwidth]{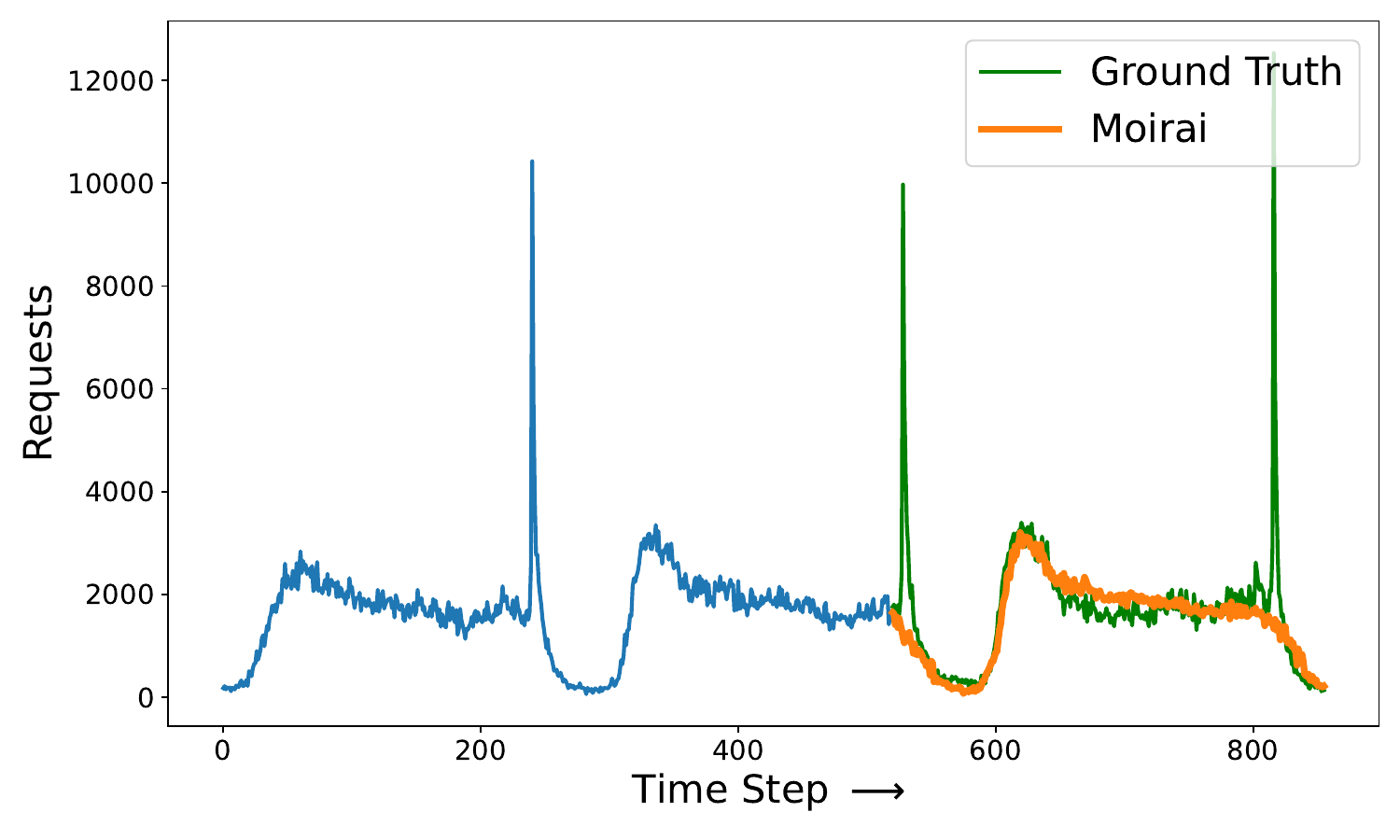}
    \end{minipage}
    \hfill
    \begin{minipage}[b]{0.49\textwidth}
        \centering
        \includegraphics[width=\textwidth]{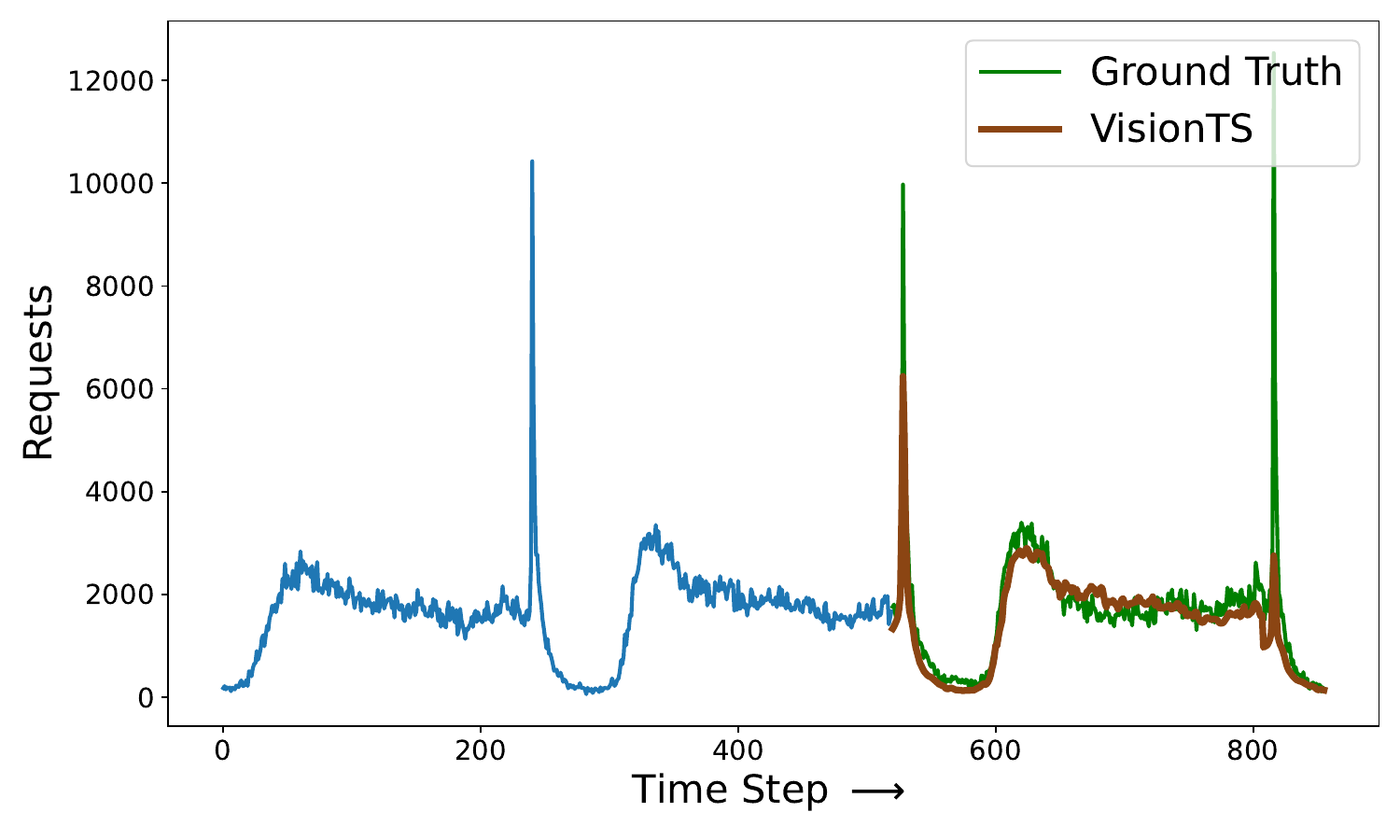}
    \end{minipage}
    \begin{minipage}[b]{0.49\textwidth}
        \centering
        \includegraphics[width=\textwidth]{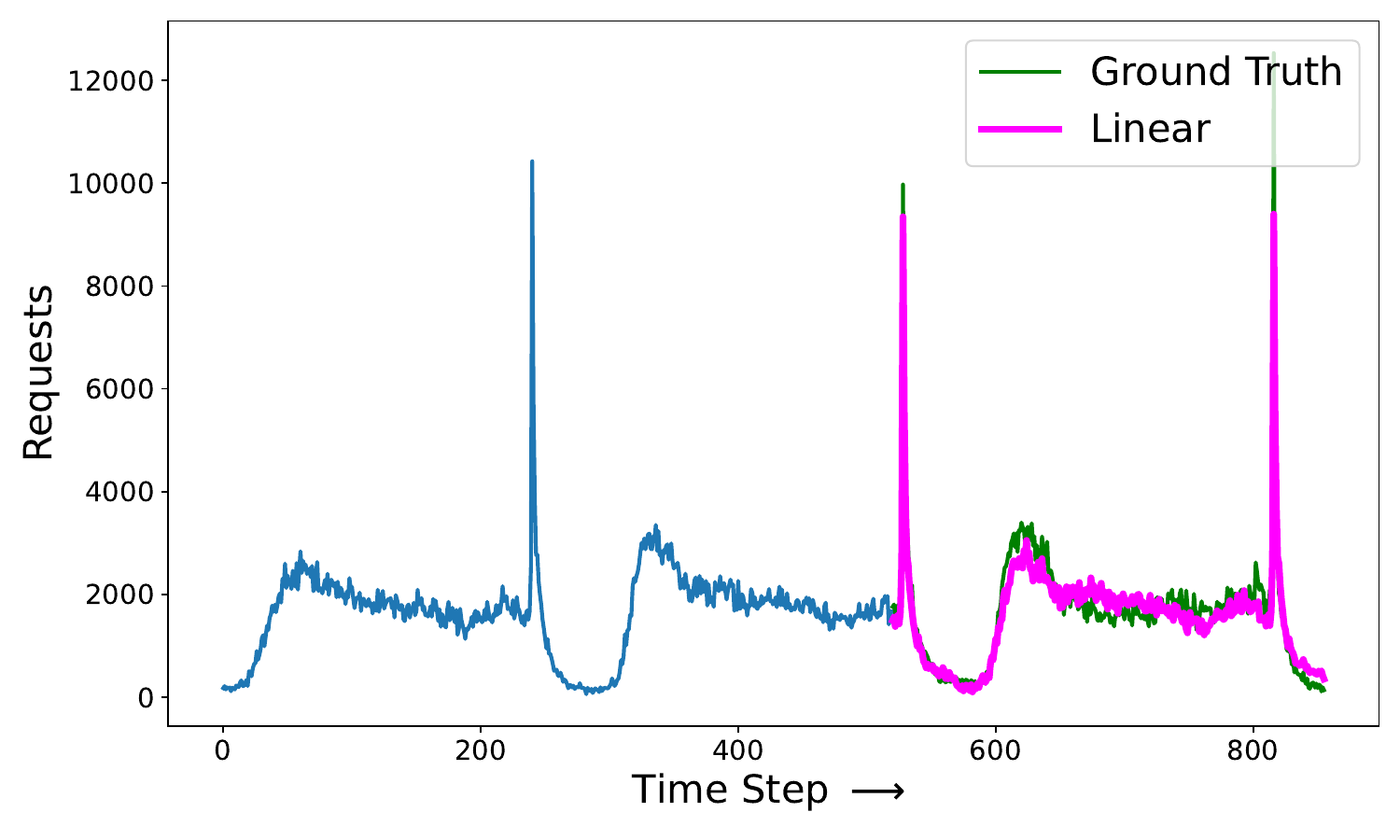}
    \end{minipage}
    \hfill
    \begin{minipage}[b]{0.49\textwidth}
        \centering
        \includegraphics[width=\textwidth]{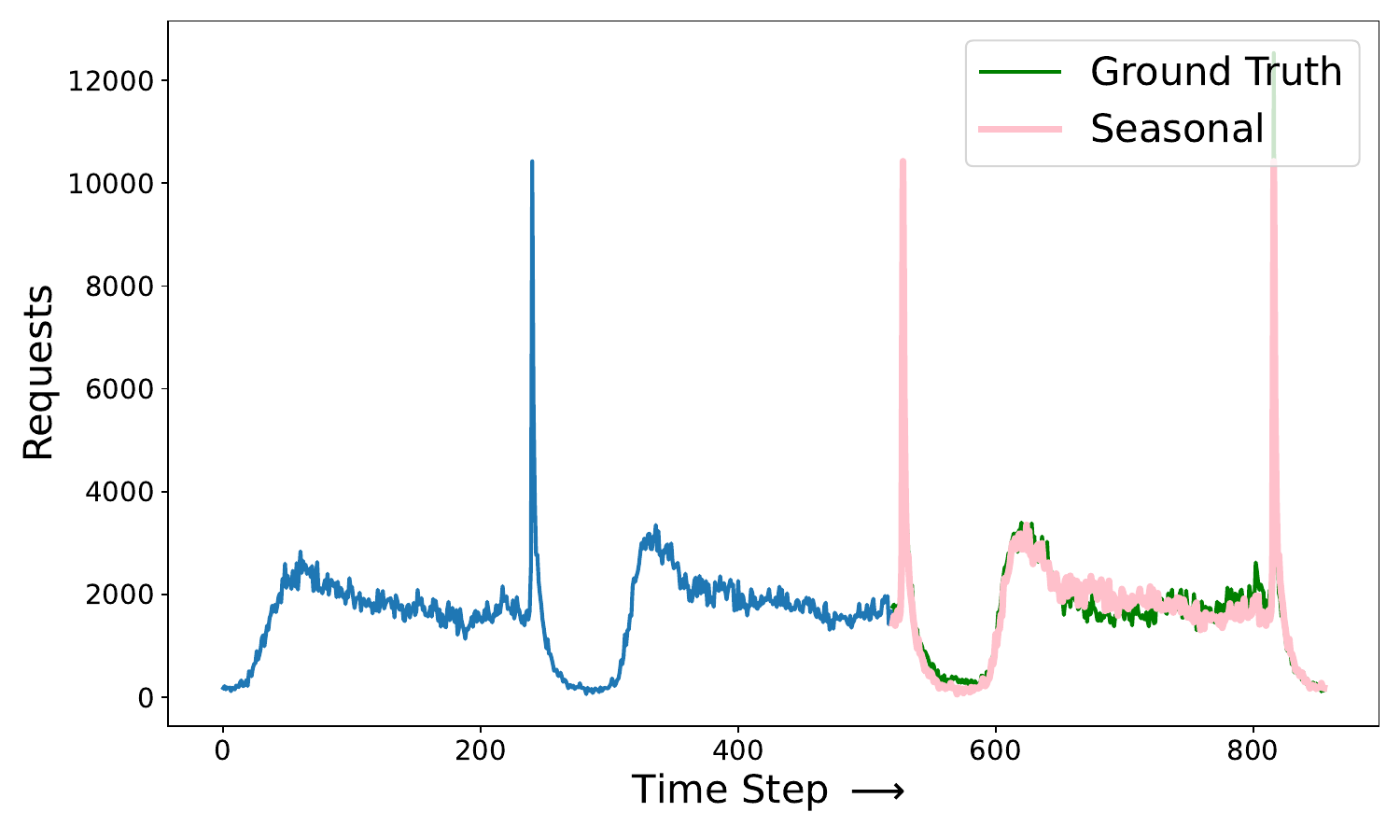}
    \end{minipage}
    \caption{\textbf{FM forecasts at the same time-step ($\bm{t=6000}$) for the third channel of the Huawei Cloud D2 dataset.} The figure shows an additional set of forecasts for the FMs and the baselines. These forecasts occur $30$ time steps before the ones shown in Figure~\ref{fig:FM_forecasts}. The figure shows that the FMs suffer from the same failures as shown in Figure~\ref{fig:FM_forecasts}, giving evidence that these problems occur frequently. None of the FMs perform as well as the simple baselines.}
    \label{fig:extra_forecasts}
\end{figure}

\newpage
\FloatBarrier
\subsection{RMSSE Results}\label{appen:rmsse}
\begin{table*}[h]
\setlength\tabcolsep{2.5pt}
\centering
\caption{\textbf{RMSSE results of zero-shot FMs forecasts for cloud data compared to the baselines of an online linear model and a naive seasonal forecaster.} The table shows, as for MASE, that the baseline methods perform the best across all datasets and forecast horizons ($H$), demonstrating that currently zero-shot FMs struggle to perform well for cloud data.}
\label{table:rmsse_results}
\begin{tabular}{@{}cc|cccccccc@{}}
\toprule
\addlinespace
& \multicolumn{1}{c}{} & & & \multicolumn{6}{c}{Zero-Shot FMs} \\
\cmidrule(l){5-10}
\multirow{2}{*}{Dataset} &  \multirow{2}{*}{$H$}  & \multirow{2}{*}{\textbf{Linear}} & \multirow{2}{*}{\textbf{Seasonal}} & \multirow{2}{*}{\textbf{VisionTS}} & \multirow{2}{*}{\textbf{TTM}} & \multirow{2}{*}{\textbf{TimesFM}} & \multirow{2}{*}{\textbf{Chronos}} & \multirow{2}{*}{\textbf{Moirai}} & \multirow{2}{*}{\textbf{Mamba4Cast}} \\ 
& & \\
\midrule
\multirow{3}{*}{D1} & 30 & \textbf{1.061} & 1.577 & 1.756 & 1.596 & 3.020 & 2.457 & 1.972 & 2.837  \\
& 96 & 1.583 & \textbf{1.557} & 1.596 & 1.814 & 4.664 & 4.551 & 2.188 & 4.419 \\
& 336 & 2.627 & \textbf{2.433} & 2.463 & 2.905 & 5.990 & 6.685 & 3.145 & 5.950 \\
\addlinespace
\multirow{3}{*}{D2} & 30 & \textbf{0.977} & 1.017 & 1.599 & 2.001 & 2.722 & 2.509 & 2.486 & 3.006 \\
& 96 & 1.358 & \textbf{1.258} & 1.927 & 2.782 & 4.819 & 4.398 & 3.782 & 6.137 \\
& 336 & 2.233 & \textbf{1.908} & 2.789 & 3.927 & 6.416 & 6.114 & 4.699 & 10.569 \\
\addlinespace
\multirow{3}{*}{D3} & 30 & \textbf{1.012} & 1.098 & 1.380 & 1.608 & 1.968 & 1.841 & 1.995 & 2.108 \\
& 96 & \textbf{1.200} & 1.237 & 1.473 & 1.971 & 3.232 & 2.757 & 2.626 & 3.481 \\
& 336 & 1.484 & \textbf{1.457} & 1.660 & 2.374 & 4.003 & 3.602 & 2.908 & 4.526 \\
\addlinespace
\multirow{3}{*}{D4} & 30 & 1.081 & \textbf{1.053} & 1.181 & 1.816 & 2.015 & 2.095 & 2.019 & 2.166 \\
& 96 & 1.194 & \textbf{1.175} & 1.233 & 1.978 & 2.410 & 2.474 & 2.321 & 2.654 \\
& 336 & 1.400 & \textbf{1.328} & 1.414 & 2.188 & 2.693 & 2.837 & 2.595 & 3.185 \\
\addlinespace
\bottomrule
\end{tabular}
\end{table*}

\end{document}